\RequirePackage{fix-cm}
\documentclass{article}

\usepackage{microtype}
\usepackage{graphicx}
\usepackage{booktabs}
\usepackage{amsmath}
\usepackage{amssymb}
\usepackage{array}
\usepackage[preprint]{icml2026}
\usepackage{hyperref}
\hypersetup{
  pdftitle={When Data Selection Pays for Itself: Fixed-Budget Value, Evidence, and What to Test Next},
  pdfauthor={Yuxiang Luo, Zhengxuan Zhang, Chen Wang, Zhuowen Liang, Xiaotian Lin, Zechao Chen, Nan Tang},
  pdfsubject={},
  hidelinks
}
\usepackage{longtable}

\icmltitlerunning{When Data Selection Pays for Itself: Fixed-Budget Value, Evidence, and What to Test Next}

\makeatletter
\renewcommand{\@pa}[1]{%
  \ifcsname the@affil#1\endcsname
  \else
    \ifcsname @icmlsymbol#1\endcsname
    \else
      \stepcounter{@affiliationcounter}%
      \newcounter{@affil#1}%
      \setcounter{@affil#1}{\value{@affiliationcounter}}%
    \fi
  \fi
  \ifcsname @icmlsymbol#1\endcsname
    \textsuperscript{\csname @icmlsymbol#1\endcsname}%
  \else
    \textsuperscript{\arabic{@affil#1}}%
  \fi
}
\renewcommand{\icmlauthor}[2]{%
  \ificmlshowauthors
    \mbox{\bf #1}\,\@for\theaffil:=#2\do{\@pa{\theaffil}}\addtofullauthorlist{#1}%
  \else
    \ifdefined\@icmlfirsttime\else
      \gdef\@icmlfirsttime{1}
      \mbox{\bf Anonymous Authors}\@pa{@anon}\addtofullauthorlist{Anonymous Authors}%
    \fi
  \fi
}
\renewcommand{\Notice@String}{}
\renewcommand{\ICML@preprint}{}
\renewcommand{\printAffiliationsAndNotice}[1]{%
  \global\icml@noticeprintedtrue
  {\let\thefootnote\relax\footnotetext{\hspace*{-\footnotesep}%
    \textsuperscript{1}\,The Hong Kong University of Science and Technology (Guangzhou), Guangzhou, China. Correspondence to: Nan Tang \textless{}nantang@hkust-gz.edu.cn\textgreater{}.}}%
}
\makeatother

\begin{document}

\twocolumn[
  \icmltitle{When Data Selection Pays for Itself:\texorpdfstring{\\}{ }
    Fixed-Budget Value, Evidence, and What to Test Next}

  \begin{icmlauthorlist}
    \icmlauthor{Yuxiang Luo}{gz}\hspace{1.15em}%
    \icmlauthor{Zhengxuan Zhang}{gz}\hspace{1.15em}%
    \icmlauthor{Chen Wang}{gz}\\
    \icmlauthor{Zhuowen Liang}{gz}\hspace{1.15em}%
    \icmlauthor{Xiaotian Lin}{gz}\hspace{1.15em}%
    \icmlauthor{Zechao Chen}{gz}\hspace{1.15em}%
    \icmlauthor{Nan Tang}{gz}%
  \end{icmlauthorlist}

  \icmlaffiliation{gz}{The Hong Kong University of Science and Technology (Guangzhou), Guangzhou, China}
  \icmlcorrespondingauthor{Nan Tang}{nantang@hkust-gz.edu.cn}

  \vskip 0.3in
]

\printAffiliationsAndNotice{}

\begin{abstract}
Selection already improves the data a model trains on, and the improvements are not one kind of gain.
Deduplication and coverage, quality and diversity, and target relevance remove different examples, pay for the decision in different places, and leave different objects to reuse.
A retained fraction of the pool does not say which of those gains was bought.
Once the cost of choosing the data shares a budget with training, the same budget also changes which data are worth choosing, so subtracting a selector's cost from a subset chosen in advance is only one of the decisions.
Deduplication that counts its overhead, compute-constrained comparisons, and selectors that update during training already locate pieces of this map.
The practical value of a plan is the performance it attains under a stated total budget and a stated reuse condition: where the gain comes from, where the cost is paid, what can be reused, and which change of model, target, pool, or stage triggers a re-check.
\end{abstract}

\section{Introduction}
\label{sec:intro}

The pool a run could train on is larger than the budget that will be spent.
A selector spends compute to find a subset, and that spend is a real part of the run whenever selection and training draw on the same budget.
Two questions then separate.
Holding training compute fixed asks whether the selected data are better data.
Asking whether to adopt the whole procedure asks whether the model is better at a given total, after the selector has been paid.
The first can be answered in the affirmative while the second goes the other way, because the compute that found the subset is compute the run did not train on.
The question is when a selection plan improves attainable performance, and one subset's success is not a general recommendation.

The gains already on record are different kinds of improvement, and the costs already on record are different kinds of payment.
Semantic deduplication and coverage, quality and diversity, and target relevance each change the training distribution, and each leaves a different object behind: a cluster assignment, a quality score, or a gradient feature tied to a model and a target.
Some of those objects are built once, offline.
Some are queried again for each task.
Some are refreshed on a schedule.
Some are recomputed inside the training loop and share their compute with the update they inform.
Section~\ref{sec:reuse} names, for a cited plan, the gain, where the cost was paid, what can be reused, and which change of model, target, pool, or stage is a reason to re-check whether that reuse still pays.

The budget already changes the decision in published comparisons, and it changes the decision in more than one way.
Under a joint budget, an expensive utility can lose to a cheaper one, and the same utility can lead when only training compute is matched.
A selector can repay its extra cost only after the selected set is large.
Separately, the data worth selecting can themselves change with the budget: how much data, which quality, and which distribution.
The last of these is why ``choose a better subset, then subtract the selector's cost'' is not the whole story.

\section{What selection already improves}
\label{sec:gains}

The value of selecting data is already visible, and the gains are different kinds.
Deduplication and coverage remove near-copies and rebalance what remains.
SemDeDup does this by embedding similarity, and on LAION the deduplication sits inside about one percent of training GPU hours, after which a large cut in the data is reported as an overall saving; its language plots are savings in training FLOPs, which is a different total from that LAION account~\citep{abbas2023semdedup}.
D4 deduplicates and diversifies documents for language-model pretraining.
It is both a mechanism and a cost.
At matched token counts the deduplicated mixture trains a better model.
That comparison is separate from the hour accounting.
In \S4.3, after deduplication and diversification, about 20\% fewer update steps reach the baseline perplexity; that step saving converts to about 4300 GPU hours before an embedding cost of about 888 GPU hours, and about 3412 GPU hours net.
A count of those update steps that ignores the embedding is the wrong efficiency.
Accuracy gains are up to about two points across 16 tasks for a 6.7B model~\citep{tirumala2023d4}.
The object left behind is an embedding index and a cluster assignment, and the cut fails when distinct documents have been merged or when a later task needed the copies that were removed.

Quality and diversity are a second gain.
QuRating scores documents with a rater fit to pairwise judgments and trains on slices of that ranking; the top of the ranking is not always the best training slice, which is the diversity sitting inside a quality score~\citep{wettig2024qurating}.
DCLM compares quality-filtered pools at matched training tokens, and the filtering step is outside that token match~\citep{li2024datacomplm}.
DEITA scores instruction examples for complexity, quality, and diversity and trains on a small selected set; that set is a complexity-and-quality cut of an instruction pool, not a deduplicated pretraining corpus~\citep{liu2024deita}.
The object left behind is a quality score, a filtered corpus, or a scored instruction set.
The failure is a high score on text the model did not need, or a filter that removed a tail the downstream tasks use.

Target relevance and transfer are a third gain.
DSIR resamples the pool toward a target distribution, so the selected data are the data that look like a chosen target rather than the data that are globally non-redundant or highly rated~\citep{xie2023dsir}.
LESS scores candidates by the alignment of their gradient features with a target validation gradient, and the headline comparison is a small fraction of the pool against the full set, with the GPU hours of building the datastore reported beside that comparison rather than inside it.
The same datastore, computed with a smaller model, can be reused to choose the subset for a larger model, which is the cross-model reuse LESS already studies~\citep{xia2024less}.
The object left behind is DSIR's target distribution, or, for LESS, a per-example gradient and a target score.

A retained fraction does not name the gain: the same ratio of the pool can be a deduplicated corpus, a quality filter, or a target-conditioned subset.
Those three pay different scoring costs, leave different objects to reuse, and fail for different reasons.

The dilemma is which of those costs is worth paying.
The following comparison is hypothetical.
Two plans are each given 100 units of training compute.
The selector's data reach a higher score, and building the subset cost 20 units beyond that training.
That result is the subset's value at that training setting.
If the real budget is 100, the selector has 80 left for training and random still has about 100.
The matched-training result does not say which plan leads after the cut.
If the missing 20 units still move the score, random can overtake a subset that was better only when both sides trained the same amount.
If those units no longer move the score, the lead can survive.
The decision is
\begin{equation}
\begin{aligned}
\max_{A,S,h}\quad
  & P\bigl(M(S,h), T\bigr) \\
\text{subject to}\quad
  & C_{\mathrm{selection}}(A)
    + C_{\mathrm{training}}(S,h) \\
  & + C_{\mathrm{maintenance}}
  \le B,
\end{aligned}
\label{eq:decision}
\end{equation}
$A$ is the selection procedure, $S$ a subset produced by $A$ from the candidate pool, $h$ the training schedule, $M(S,h)$ the model trained on that subset under that schedule, and $T$ the target.
Compute shared by selection and training is counted once in the sum, as in Figure~\ref{fig:controls}.
The selection term may be an offline index, a per-task query, a periodic rescore, or compute spent inside the training loop.
Adding an example costs more training compute when the schedule is a fixed number of epochs, and it can displace other tokens when the schedule is a fixed token budget.
The schedule is part of which price is being paid.

Matched training compute asks whether the data are better when every plan trains the same amount.
A shared total asks whether to run the procedure, because the selector is paid from the budget the training would have used, and each plan then spends what remains.

\section{What is already known}
\label{sec:known}

Train-only comparisons and total-budget comparisons are already both in the literature, and they disagree in a regular way.
Yin and Rush put the cost of the utility and the cost of training on one budget.
Under that budget, cheaper lexicon and embedding utilities dominate at small and medium budgets, and perplexity and gradient utilities join the compute-optimal frontier only at training-to-selection model-size ratios of about $5\times$ and about $10\times$, ratios obtained as parametric fits extrapolated from their smaller models and checked in that finetuning setting.
Hold the training budget fixed, and the expensive utilities lead~\citep{yin2025ccds}.
Ivison et al.\ keep the same separation in another experimental bed.
Selectors compared at matched sample counts answer which subset is better data at that count.
A separate curve charges selection against balanced random, and RDS+ repays the extra cost only once the selected set reaches roughly 326k examples.
That figure is the size of the selected set on the curve~\citep{ivison2025largescale}.
D4, as above, is the pretraining counterpart: the matched-token comparison is separate from the \S4.3 conversion of fewer update steps into GPU hours.
The shared question is a stated procedure, a cost with a place, and a recommendation that moves with scale.

Sorscher et al.\ give a theoretical argument and vision experiments in which pruning the training set can beat a power-law scaling curve~\citep{sorscher2022beyond}.

The budget also changes which data are worth selecting.
CADS treats the compute budget as an input to selection rather than as a bill paid after a subset has been chosen.
Different budgets call for different quantities, different qualities, and different distributions, and the method is a bilevel procedure whose inner loop trains under the budget on the subset the outer loop is choosing~\citep{wan2025cads}.
That is the sense in which choosing better data and then subtracting the selector's cost is incomplete.
The subtraction answers whether a subset already in hand was worth the cost of its score.
It does not answer whether a larger budget should have selected different data.
CCDS and Ivison et al.\ compare procedures under a budget.
CADS lets the selected distribution move with the budget.

Reuse and refresh are already part of the same literature.
LESS reuses a gradient datastore built with a smaller model when the model to be trained is a larger one, which is transfer of the selection rather than a new datastore per model~\citep{xia2024less}.
Yin and Rush spread the gradient cost across target tasks that share a pool and a selection model.
With Llama-2 7B as that model, the joint-budget performance reaches their finetuned Pareto frontier at 10 tasks and passes it at 20~\citep{yin2025ccds}.
Refreshing Less starts from the other end of the same pipeline.
Gradient features cached at a post-warmup checkpoint, paired with a fresh validation gradient, still rank candidates 40 optimizer steps later at Spearman correlation between 0.952 and 0.991, while the top tenth they induce misses 10 to 22 percent of the examples a full recomputation would have kept.
The method refreshes gradient features only for a top-ranked fraction, calibrates the stale scores on the recomputed examples, and uses the agreement between stale and recomputed scores on those examples as a check on the reuse.
At a refresh fraction of 0.3 the recovered top-$k$ matches full recomputation in their settings, and the gradient stage is several times cheaper~\citep{su2026refreshing}.

Selection also already updates during training and shares its compute with the training step.
MATES fits an influence model on the run in progress and reselects as pretraining proceeds.
Selection is 21.7 percent of total FLOPs at 410M and 11.5 percent at 1B on a 25B-token run, and at 1B a matched downstream performance costs 43.3 percent of the random run's FLOPs, selection included~\citep{yu2024mates}.
LEAD estimates each example's utility inside the ordinary training loop, from the current loss, a gradient estimate of how the loss will change, and a smoothed history, and it chooses the next subset without a separate pass over the whole pool~\citep{lin2025lead}.

\section{Interpreting the existing evidence}
\label{sec:claims}

Studies that already count selection inside the cost still cannot be stacked into one threshold or one method ranking.
Two axes stay distinct.
One is whether the selector depends explicitly on the total budget of the run: a scorer with fixed parameters can still change how much of the pool it keeps as that budget changes, and a selector can also change which examples it keeps.
The other is whether the selector updates with the model state during a run.
The total budget of a run and the stage inside one run are different variables, and a result on one axis is not a measurement of the other.

\begin{figure*}[t]
  \centering
  \includegraphics[width=\textwidth]{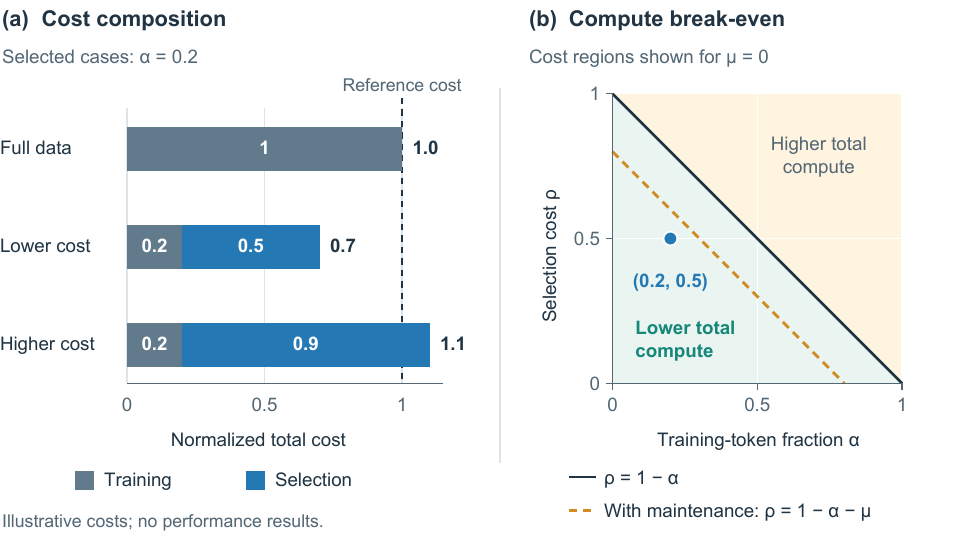}
  \caption{Cost split for a fixed model whose training cost is linear in tokens processed.
  (a)~Full-data training is normalized to one.
  The drawn cases use $\alpha=0.2$ with $\rho=0.5$ and with $\rho=0.9$, totals $0.7$ and $1.1$, excluding maintenance.
  $\alpha$ is the share of training tokens actually processed, including repeats, and $\rho$ is selection cost over a full training run.
  (b)~Whether total cost is below one full training run.
  The shaded region is that lower total when maintenance is left out, and a random subset falls in it as well.
  A maintenance share $\mu>0$ moves the cost boundary to $\rho=1-\alpha-\mu$ (dashed).
  Scores at a given total are a separate comparison.}
  \label{fig:cost}
\end{figure*}

Equation~\eqref{eq:breakeven} and Figure~\ref{fig:cost} are the cost half of reading any of these totals.
For a fixed model whose training cost is approximately linear in tokens processed, let $\alpha$ be the share of training tokens actually processed, repeats included, and let $\rho$ be selection cost divided by one full training run.
The selector uses less compute than one full pass when
\begin{equation}
  \rho < 1-\alpha.
  \label{eq:breakeven}
\end{equation}
A maintenance share $\mu$ moves that cost boundary to $\rho = 1-\alpha-\mu$.
The inequality compares cost with a full run.
The same total can train a better model or a worse one.
Performance break-even is a comparison of scores at one shared total, selection included, and it is a different object from the curve in the figure.
A random subset with almost no selection cost meets the cost inequality and still has its own score.
Beating that random arm is a comparison with random.
A frontier over cheaper selectors is a further comparison among those selectors at the same total.

The size of the score gap can move for more than one reason, and each reason is a hypothesis about a stated comparison.
A flat training curve is the hypothesis that additional tokens barely move the score, so compute spent on selection subtracts little from a matched-training lead; on a steep part of the curve the same spend subtracts more.
Redundancy is the hypothesis that many candidates are interchangeable for this model and this target.
A subset that beats a random subset at matched training compute is consistent with redundancy and also with a difference in quality or relevance, so the win does not by itself identify the gain as deduplication.
A matched signal is the hypothesis that candidates were scored for the target being trained.
A mechanism ablation matches the training budget and changes one of those factors.
Factors that are not controlled or matched may still confound the interpretation.
A practical-value comparison matches the total budget and lets each plan spend its remainder.
Requiring the same total FLOPs and the same number of training tokens leaves leftover compute, and the comparison has to say whether that remainder is spent, discarded, or charged elsewhere.
The MATES sections report the selection share of FLOPs and the total at matched performance; they leave these hypotheses open.

\section{Where the cost is paid}
\label{sec:reuse}

Where the gain comes from is coverage, quality, or target relevance, as in Section~\ref{sec:gains}.
Where the cost is paid is an offline index, a per-task query, a periodic rescore, or incremental compute during training.
MATES and LEAD are the incremental case: one updates an influence model during pretraining, and the other estimates utility inside the training step.
A cost analysis written only as keep, rescore, or rebuild has no term for that case.

An embedding index is reused across targets and models on the pool that was embedded, and a new pool is the trigger.
A quality score is reused across models, and a new notion of quality is the trigger.
A per-example gradient is reused across tasks on a fixed model, the CCDS amortization in Section~\ref{sec:known}, and across model scale, where LESS uses a smaller model's datastore to select for a larger model; a new model, target, or stage is the trigger, and that LESS result is a case in which the re-check keeps the object.
A stale gradient cache is reused for a limited run of optimizer steps and then partially refreshed, as Refreshing Less does, and a move in training stage is the trigger.
The final subset is reused while the ranking that produced it still fits the run, and a change of model, target, pool, or stage is the trigger.
Mixture weights are included with subsets: DoReMi learns them on a small proxy and reuses them for a larger model, the cost is paid on the proxy, and a new pool or target is the trigger~\citep{xie2023doremi}.

Reuse pays when the skipped recomputation was real and the kept object still carries a selection gain.
A cheap cache of a bad subset saves the wrong quantity.
A faithful index that loses to spending the same budget on training is a poor purchase under the total-budget comparison.

\section{What to compare next}
\label{sec:next}

\begin{figure*}[t]
  \centering
  \includegraphics[width=\textwidth]{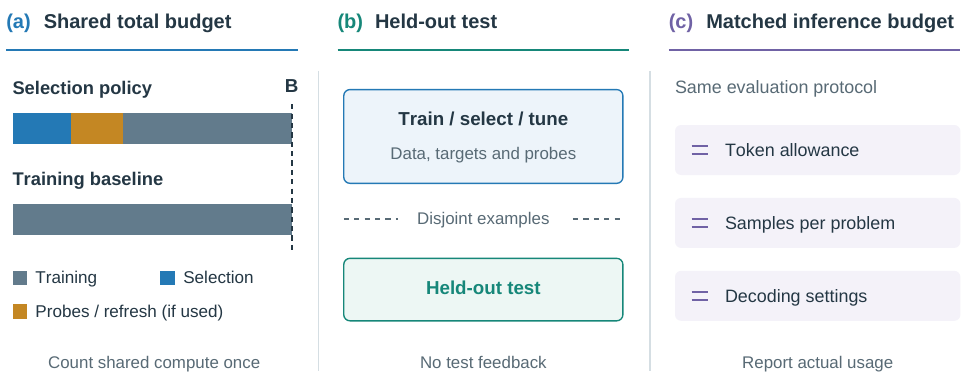}
  \caption{Evaluation controls.
  (a)~Shared total budget: compare scores at the same total, with probe and refresh costs inside that total when they are used.
  Compute shared by training and selection is counted once.
  Bar allocations are schematic.
  (b)~Held-out test: keep final test examples out of training, selection, and tuning, with no test feedback.
  (c)~Matched inference budget: match the inference allowance and the decoding settings, and report actual usage.}
  \label{fig:controls}
\end{figure*}

Figure~\ref{fig:controls} is the protocol: the test is held out of scoring, tuning, and training, the inference budget matches, and training and selection variation are reported with the uncertainty of the difference.
These comparisons match the total budget and let each plan spend its remainder.
Added training is new data from the pool, another pass over data already seen, or a longer run on the selected set.
Extra epochs on repeated data have diminishing value, so each comparison names which of the three the remainder buys~\citep{muennighoff2023dataconstrained}.
When the gain is deduplication, score targets stay out of the test, the pool is checked for test contamination, and a threshold tuned on a validation set is reported with the cost of that tuning and the bias of the choice~\citep{lee2022dedup}.

\paragraph{Refresh against more training.}
Section~\ref{sec:known} already scores a stale gradient cache and a partial refresh against full recomputation and against random, not against more training.
The comparison asks whether agreement between stale and recomputed scores, or the top-$k$ overlap measured with it, ranks the net gain of refreshing versus a named form of extra training, across states of the cache.
The prediction for that comparison is a ranking of that net gain, not a sign: lower agreement corresponds to a larger net gain.
The criterion is a reversal of that ranking across states.
The probe that decides whether to refresh is charged before that decision; agreement on examples already refreshed describes the refresh that was run.
TuneAhead's probe is a separate pre-finetune check: static dataset descriptors plus a short probe give a go or no-go before full training~\citep{luo2026tuneahead}.
Luo, Wang, and Tang decompose Bayes-optimal prediction risk into a floor, set by static data--model mismatch and task stochasticity, which does not fall as the probe lengthens, and an optimization-variance term, which does~\citep{luo2026risk}.
Further probe budget buys the second term, and it changes the decision while that term is still large enough to move the go or no-go.

\paragraph{An incremental update against a full rescore.}
Section~\ref{sec:known} already charges an in-training update to the run, through MATES's influence model and through LEAD's utility computed from the training step.
The comparison is that update against a periodic full rescore of the same pool and the same gain, at one shared total, with the displaced training named as new tokens, a repeated pass, or a longer run on the current subset.
What is compared is time since the last rescore: a full rescore updates the scores, so that interval is the cache age, and distance from the first checkpoint is not.
The question is how the net advantage varies with cache age at a fixed total budget.
Each paired comparison holds the total budget of the run fixed.

\paragraph{Budget dependence against a fixed-parameter scorer.}
Section~\ref{sec:known} already separates procedures whose cost is counted at a given scale from CADS, whose rule takes the budget as an input and changes how much data, which quality, and which distribution are kept.
Dependence on the total budget and updating with model state are different axes.
The second is the comparison above.
MATES updates with model state; the fixed arm here is a scorer whose parameters do not take the budget as an input.
This comparison holds the scorer parameters fixed on one arm.
That arm can still change its retention rate as the total budget changes.
The other arm takes the total budget as an input and can change which examples are kept.
Both are scored at the same totals, the stage inside each run is not the variable being swept, and the remainder is named as new data, repeats, or a longer run.
The roughly 326k point in Section~\ref{sec:known} is a selected-set size on a FLOPs curve.
The prediction asks over which budget range the budget-dependent rule has a net gain over the fixed-parameter scorer, and whether that gain comes from a change in the data distribution beyond the change in retention rate.
Subset overlap can explain a difference in the sets; the performance difference at the shared total decides the practical value.

\section{Discussion}
\label{sec:discuss}

A fixed training budget reads which data are better when every plan trains the same amount, and a shared total reads whether to run the selector, since the two rankings can disagree as in Section~\ref{sec:known}.

A public index changes who pays the offline cost.
The user who downloads a released embedding index, a released quality score, or a released gradient datastore decides at the margin: apply it, refresh part of it, or ignore it.
That marginal decision does not pay the build.
The system-level question is whether the build improved the frontier across the uses that share it by enough to cover the builder's cost, which depends on how many of those uses the index still fits.
We answer the question of the party whose own budget pays for the selection on the runs being trained, which is the record in Section~\ref{sec:reuse} applied to those runs.

\section{Conclusion}
\label{sec:conclusion}

The practical value of a selection plan is the performance it attains under a stated total budget and a stated reuse condition.
The four-part frame changes a citation from a retained fraction into a record of where the gain comes from, where the cost is paid, what object can be reused, and which change of model, target, pool, or stage triggers a re-check.
With that record, a saving after deduplication, a compute-optimal utility, and a budget-dependent distribution remain different facts, and a shift in scale or in model state can change the recommendation while the stored object is still worth checking.
Report the full cost inside the budget, and compare selection, reuse, or refresh with the other uses that budget could have bought.

\bibliographystyle{icml2026}
\bibliography{refs}

\onecolumn
\appendix
\section{Evidence for the argument}
\label{app:notes}

Table~\ref{tab:evidence} lists the papers the body uses and the role each comparison plays.
Table~\ref{tab:notes} records the source of the figures cited above.

\begingroup
\fontsize{7.5pt}{8.6pt}\selectfont
\setlength{\tabcolsep}{3pt}
\renewcommand{\arraystretch}{1.0}
\begin{longtable}{@{}p{0.28\columnwidth}p{0.33\columnwidth}p{0.33\columnwidth}@{}}
  \caption{Comparisons used in the body, grouped by the claim they support.}
  \label{tab:evidence}\\
  \toprule
  \raggedright\arraybackslash Work &
  \raggedright\arraybackslash What is on record &
  \raggedright\arraybackslash Role in the argument \\
  \midrule
  \endfirsthead
  \toprule
  \raggedright\arraybackslash Work &
  \raggedright\arraybackslash What is on record &
  \raggedright\arraybackslash Role in the argument \\
  \midrule
  \endhead
  \bottomrule
  \endfoot
  \raggedright\arraybackslash SemDeDup~\citep{abbas2023semdedup} &
  \raggedright\arraybackslash Semantic deduplication. On LAION the overhead is about one percent of training GPU hours and a large cut is then an overall saving. Language plots are training-FLOP savings. &
  \raggedright\arraybackslash Coverage gain. The LAION total and the language FLOP plots are different accounts. \\
  \raggedright\arraybackslash D4~\citep{tirumala2023d4} &
  \raggedright\arraybackslash Deduplication and diversification. \S4.1 matches tokens. \S4.3 converts about 20\% fewer update steps to baseline perplexity into about 4300 GPU hours before an embedding cost of about 888, net about 3412. Up to about two points on 16 tasks at 6.7B. &
  \raggedright\arraybackslash Mechanism and cost together. \\
  \raggedright\arraybackslash DEITA~\citep{liu2024deita} &
  \raggedright\arraybackslash Instruction examples scored for complexity, quality, and diversity, then a small selected set. &
  \raggedright\arraybackslash A complexity-and-quality cut of an instruction pool, distinct from a deduplicated pretraining corpus. \\
  \raggedright\arraybackslash Lee et al.~\citep{lee2022dedup} &
  \raggedright\arraybackslash Deduplication reduces memorization and train--test overlap. &
  \raggedright\arraybackslash Connects a coverage gain to the held-out protocol: targets versus test, pool contamination, and the cost and bias of a threshold tuned on a validation set. \\
  \raggedright\arraybackslash Sorscher et al.~\citep{sorscher2022beyond} &
  \raggedright\arraybackslash A theoretical argument and vision experiments in which pruning can beat a power-law scaling curve. &
  \raggedright\arraybackslash A source for pruning and scaling in that theory and those experiments. \\
  \raggedright\arraybackslash DoReMi~\citep{xie2023doremi} &
  \raggedright\arraybackslash Mixture weights learned on a small proxy and reused for a larger model. &
  \raggedright\arraybackslash Weights as a reusable object in the same record as subsets. \\
  \raggedright\arraybackslash Muennighoff et al.~\citep{muennighoff2023dataconstrained} &
  \raggedright\arraybackslash Extra epochs on repeated data have diminishing value. &
  \raggedright\arraybackslash Added training has to say whether it is new data, a repeated pass, or a longer run. \\
  \raggedright\arraybackslash QuRating~\citep{wettig2024qurating} &
  \raggedright\arraybackslash Quality ratings. The top of the ranking is not always the best training slice. Rater cost is outside the training comparison. &
  \raggedright\arraybackslash Quality and diversity. Evidence about the data until the rater is inside the total. \\
  \raggedright\arraybackslash DCLM~\citep{li2024datacomplm} &
  \raggedright\arraybackslash Quality-filtered pools at matched training tokens. Filtering is outside that match. &
  \raggedright\arraybackslash Quality under a fixed training plan. \\
  \raggedright\arraybackslash DSIR~\citep{xie2023dsir} &
  \raggedright\arraybackslash Importance resampling toward a target distribution. &
  \raggedright\arraybackslash Target relevance. The reuse object is the target distribution. \\
  \raggedright\arraybackslash LESS~\citep{xia2024less} &
  \raggedright\arraybackslash Target-gradient alignment. A small fraction of the pool against the full set, selection GPU hours beside that comparison. A smaller model's datastore can select for a larger model. &
  \raggedright\arraybackslash Target relevance, and reuse across model scale. \\
  \raggedright\arraybackslash CCDS~\citep{yin2025ccds} &
  \raggedright\arraybackslash Joint budget. Cheaper utilities win at small and medium budgets. Perplexity and gradient selection join at about $5\times$ and $10\times$, as fits in that setting. Fixed training budget reverses the ranking. Gradient cost across tasks reaches the frontier at 10 tasks and passes it at 20 with Llama-2 7B selecting. &
  \raggedright\arraybackslash Train-only versus total. Cross-task reuse on a fixed model and pool. \\
  \raggedright\arraybackslash Ivison et al.~\citep{ivison2025largescale} &
  \raggedright\arraybackslash Matched sample counts, and a FLOPs curve that includes selection. RDS+ repays its extra cost from roughly 326k selected examples. &
  \raggedright\arraybackslash Scale along selected-set size. That figure is not a candidate-pool size. \\
  \raggedright\arraybackslash CADS~\citep{wan2025cads} &
  \raggedright\arraybackslash The budget changes how much data, which quality, and which distribution are worth selecting. &
  \raggedright\arraybackslash The rule takes the total budget as an input. Distinct from a fixed-parameter scorer and from an update with model state. \\
  \raggedright\arraybackslash MATES~\citep{yu2024mates} &
  \raggedright\arraybackslash Influence model updated during pretraining. Selection is 21.7\% of FLOPs at 410M and 11.5\% at 1B. At 1B, matched performance at 43.3\% of the random run's FLOPs. &
  \raggedright\arraybackslash Incremental cost during training. Cost share and matched-performance FLOPs. \\
  \raggedright\arraybackslash LEAD~\citep{lin2025lead} &
  \raggedright\arraybackslash Utility estimated inside the training loop, without a separate full-pool inference pass. &
  \raggedright\arraybackslash Incremental cost. The selection pass and the training pass coincide. \\
  \raggedright\arraybackslash Refreshing Less~\citep{su2026refreshing} &
  \raggedright\arraybackslash Stale gradient cache, partial refresh of a top fraction, agreement between stale and recomputed scores. Ranking largely preserved after 40 steps; top tenth misses 10--22\% of a full recomputation. &
  \raggedright\arraybackslash Reuse and refresh already built. Scored against full recomputation, not against more training. \\
  \raggedright\arraybackslash TuneAhead~\citep{luo2026tuneahead} &
  \raggedright\arraybackslash Static descriptors plus a short probe, then a go or no-go before full fine-tuning. &
  \raggedright\arraybackslash A small spend before the rest of the budget. \\
  \raggedright\arraybackslash Risk decomposition~\citep{luo2026risk} &
  \raggedright\arraybackslash Pre-hoc risk splits into a floor invariant to probe length and an optimization variance that falls as the probe lengthens. &
  \raggedright\arraybackslash Which part of a probe still moves the decision. \\
\end{longtable}

\endgroup

\begingroup
\footnotesize
\setlength{\tabcolsep}{4pt}
\renewcommand{\arraystretch}{1.05}
\begin{longtable}{@{}p{0.22\columnwidth}p{0.74\columnwidth}@{}}
  \caption{Sources for the figures in Table~\ref{tab:evidence}.}
  \label{tab:notes}\\
  \toprule
  \raggedright\arraybackslash Work &
  \raggedright\arraybackslash Source of the cited figures \\
  \midrule
  \endfirsthead
  \multicolumn{2}{@{}l}{\tablename\ \thetable, continued}\\[0.4em]
  \toprule
  \raggedright\arraybackslash Work &
  \raggedright\arraybackslash Source of the cited figures \\
  \midrule
  \endhead
  \bottomrule
  \endfoot
  \raggedright\arraybackslash SemDeDup &
  \raggedright\arraybackslash \S7, for LAION, puts deduplication within about 1\% of training GPU hours and treats a large cut as a saving after that overhead. \S5 plots training-FLOP savings on language data; that plot is not the LAION total. \\
  \raggedright\arraybackslash D4 &
  \raggedright\arraybackslash \S4.1 is a fixed-token comparison, separate from the hour accounting. \S4.3, after deduplication and diversification, reaches baseline perplexity in about 20\% fewer update steps, converted to about 4300 GPU hours before an embedding cost of about 888 GPU hours, and about 3412 GPU hours net. Accuracy gains are up to about 2\% on 16 tasks at 6.7B. \\
  \raggedright\arraybackslash DEITA &
  \raggedright\arraybackslash Instruction data are scored for complexity and quality, with a diversity step, and a small selected set is used for instruction tuning. \\
  \raggedright\arraybackslash Lee et al. &
  \raggedright\arraybackslash Deduplicating the training corpus reduces memorization of training text and reduces overlap between train and test, in addition to the effect on loss. \\
  \raggedright\arraybackslash Sorscher et al. &
  \raggedright\arraybackslash The pruning result is a theoretical argument together with vision experiments. It is the source cited for beating a power-law curve in those settings. \\
  \raggedright\arraybackslash DoReMi &
  \raggedright\arraybackslash Group mixture weights are optimized with a small proxy model and then used to train a larger model. The reusable object is the weight vector. \\
  \raggedright\arraybackslash Muennighoff et al. &
  \raggedright\arraybackslash Under a data constraint, repeating data for extra epochs returns diminishing improvements. A repeated pass, new data, and a longer run on a fixed set are different uses of added compute. \\
  \raggedright\arraybackslash QuRating &
  \raggedright\arraybackslash \S5 trains 1.3B models on 30B tokens drawn from a pool scored by a rater fit to GPT-3.5 pairwise judgments. The top of the ranking is not always best. The cost of the judgments and of the rater is outside the training comparison. \\
  \raggedright\arraybackslash DCLM &
  \raggedright\arraybackslash \S3.2 matches training tokens. Filtering cost is outside that comparison. \\
  \raggedright\arraybackslash DSIR &
  \raggedright\arraybackslash Importance weights resample the pool toward a target distribution. The selected set is defined by that target. \\
  \raggedright\arraybackslash LESS &
  \raggedright\arraybackslash The headline comparison is a small fraction of examples against the full set. Selection GPU hours are reported for the datastore and sit outside that matched-count comparison. The paper also uses a smaller model's gradient features to select data for a larger model. \\
  \raggedright\arraybackslash CCDS &
  \raggedright\arraybackslash The joint objective charges training and the utility to one budget. At small and medium budgets, lexicon and embedding utilities beat perplexity and gradient utilities. The $5\times$ and $10\times$ ratios are parametric fits extrapolated from the smaller models and checked against the finetuning runs. Figure~5a holds the training budget fixed, and the gradient and perplexity utilities then lead. Figure~4 spreads the gradient cost across tasks: with Llama-2 7B selecting, the finetuned Pareto frontier is reached at 10 tasks and passed at 20. \\
  \raggedright\arraybackslash Ivison et al. &
  \raggedright\arraybackslash \S4.3, Figure~3 plots performance against FLOPs that include selection, for RDS+ versus balanced random. The extra cost pays off from roughly 326k selected examples. \S4.1--4.2 compare other selectors at matched sample counts. \\
  \raggedright\arraybackslash CADS &
  \raggedright\arraybackslash The paper's argument and method: different budgets impose different requirements on quantity, quality, and distribution, and selection is posed as a bilevel problem whose inner loop trains under the budget. \\
  \raggedright\arraybackslash MATES &
  \raggedright\arraybackslash \S5 and Figure~3 measure FLOPs that include selection, on a 25B-token run. Selection is 21.7\% of the total at 410M and 11.5\% at 1B. At 1B the same performance as random selection is reached at 43.3\% of the total FLOPs. \\
  \raggedright\arraybackslash LEAD &
  \raggedright\arraybackslash The utility is computed from the training step, combining the current loss, a gradient estimate of the loss change, and a smoothed history, and the next subset is chosen without a separate inference pass over the pool. \\
  \raggedright\arraybackslash Refreshing Less &
  \raggedright\arraybackslash Gradient features cached after warmup, with a fresh validation gradient, keep Spearman correlation between 0.952 and 0.991 forty steps later. The induced top 10\% misses 10 to 22\% of the examples full recomputation selects. Refreshing a top fraction of 0.3 recovers that top-$k$ in the reported settings. Agreement between stale and recomputed scores on the refreshed examples is the reuse check. Downstream numbers are against full recomputation and random selection. \\
  \raggedright\arraybackslash TuneAhead &
  \raggedright\arraybackslash A candidate run is represented by static dataset descriptors plus a short standardized probe, and the prediction is used as a go or no-go before full fine-tuning. \\
  \raggedright\arraybackslash Risk decomposition &
  \raggedright\arraybackslash Proposition~4.1 writes pre-hoc risk as an intrinsic term, identified with static data--model mismatch and task stochasticity, plus optimization variance. The intrinsic term does not fall as the probe lengthens. Optimization variance does. \\
\end{longtable}

\endgroup

\end{document}